\documentclass[3p]{elsarticle}
\usepackage{booktabs}

\usepackage{lineno,hyperref}
\modulolinenumbers[5]
\usepackage{setspace}
\journal{Journal of \LaTeX Templates}
\usepackage{amsmath}
\usepackage{amssymb}
\usepackage{natbib}
\usepackage{amsfonts}
\usepackage{dsfont}
\biboptions{authoryear}
\begin{document}

\begin{frontmatter}

\title{Machine Learning for Multi-Output Regression: When should a holistic multivariate approach be preferred over separate univariate ones?}

\author[1]{Lena Schmid\corref{cor1}}
\ead{lena.schmid@tu-dortmund.de}
\author[1]{Alexander Gerharz}
\author[1]{Andreas Groll}
\author[1]{Markus Pauly}
\address[1]{Department of Statistics, TU Dortmund University, 44227 Dortmund, Germany}

\cortext[cor1]{Corresponding Author}

\begin{abstract}
Tree-based ensembles such as the Random Forest are modern classics among statistical learning methods. 
In particular, they are used for predicting univariate responses. 
In case of multiple outputs the question arises whether we separately fit univariate models or directly follow a multivariate approach. For the latter, 
several possibilities exist that are, e.g.\ based on modified splitting or stopping rules for multi-output regression. 
In this work we compare these methods in extensive simulations to help in answering the primary question when to use multivariate ensemble techniques.
\end{abstract}

  
\end{frontmatter}

\section{Introduction}

Multivariate data occur in a variety of disciplines, for example in biomedical research, the social sciences, or econometrics. Data are said to be multivariate if the response not only consists of one variable, but of $d\geq 2$ output variables, say ${\bf Y}\in \mathbb{R}^d$. Then, we are often interested in finding a functional relationship between the output ${\bf Y}$ and some feature variables ${\bf X}\in \mathbb{R}^p$, i.e.\ we want to perform a multivariate (also called multi-output) regression analysis. Unlike univariate multiple regression (with $d=1$), which also includes multiple features ${\bf X}\in \mathbb{R}^p$, multivariate regression wants to specify the relationship of several outcome variables with ${\bf X}$ simultaneously. The hope of such multivariate analyses is, that the consideration of possible dependencies between the outcomes may lead to procedures with better power (in case of inference) or accuracy (in case of prediction) compared to separate univariate analyses. While the need for the development and use of valid and distributional robust or nonparametric multivariate methods has been recognized and addressed in inferential statistic \citep{dobler2020nonparametric,friedrich2019resampling,konietschke2015parametric,smaga2017bootstrap,vallejo2012robust,zimmermann2020multivariate}, there do not exist exhausting studies that exploit the potential of multivariate regression methods for prediction.\\ 

Focussing on tree-based ensemble methods as the Random Forest, it is the aim of this manuscript to close this gap. In particular, we want to answer our research-motivating question:
\begin{center} 
{\it When should a holistic multivariate regression approach be preferred over separate univariate predictions?}
\end{center}
The answer to this question is ad hoc not clear. In fact, univariate tree-based ensembles as the Random Forest \citep{Breiman2001} or Extra Trees \citep{Geurts2006} have been shown to be good predictive tools in various applications \citep{pauly2012random,gerke2018childhood,schauberger2018predicting,groll2019hybrid,huang2020travel} and are applied far more often than multivariate regression approaches. One reason for this could be that there are still only a few multivariate extensions, such as \citep{Death2002, Larsen2004,DAMBROSIO201721, Zhang1998, Zhang2008}, where the impurity measure used in the tree construction was multivariatly extended. \citet{Zhang1998} developed an impurity function based on a generalized entropy criterion to handle multiple binary outputs, this work was extended by \cite{Zhang2008} to ordinal outputs by transforming the outputs to binary-valued indicator functions. \cite{Sicilano2000} used the weighted sum of Gini index reduction to construct
trees for multiple binary outputs. One of the first approaches for multi-output regression trees was proposed by \cite{Death2002}, which extended CART to multi-output regression by using the sum of squared errors over the multivariate outputs as the impurity function of a node (\cite{Death2002} also developed more general forms of distance-based impurity functions). This method is implemented in the R-package \texttt{MVPART} \citep{MVPART} and was used by \cite{Segal2011} to develop multivariate Random Forests. \cite{Larsen2004} suggested a multivariate regression tree that uses the Mahalanobis distance as node impurity function where the covariance matrix is estimated from the whole data set.  The Mahalanobis distance is also used in the R-packages \texttt{MultivariateRanfomForest} and \texttt{IntegratedMRF} \citep{Rahmann2017} to construct multivariate Random Forests. Furthermore, \cite{Rahmann2017} have observed in several drug response predictions that multivariate Random Forests provide higher accuracy than Random Forests when outputs are highly correlated. The Eart Mower and Mallows distances were used as impurity functions by \cite{DAMBROSIO201721} to develop regression trees for multivalued numerical outputs.There are also several approaches that extended the GUIDE algorithm \citep{Loh2002} to multivariate and longitudinal outputs \citep{Loh2013, Hsiao2007}.  \cite{Lee2005}  developed a method that can be used with multivariate outputs of any type. Here, GEE techniques are used to find the splits. 
Multivariate trees for a mixture of categorical and continuous outputs were presented by \cite{DINE2009}, where the splits are derived from a likelihood-based approach for a general location model.
For Extra Trees, there is also an extension for multi-task learning \citep{Simm2014}, where the split criterion has been adapted to handle multiple tasks. The method is implemented in the R-package \texttt{extraTrees} \citep{Simm2014}.\\

To answer the central research question, we compare the predictive accuracy of separate univariate analyses with a simultaneous multivariate analysis by means of exhaustive simulations and in an illustrative data analysis.\\ 

This work is structured as follows: Section~\ref{sec:Methods} presents the univariate and multivariate ensemble methods. More precisely, univariate and multivariate Random Forest and Extra Trees algorithms are presented. In addition, the multi-task Extra Trees algorithm \cite{Simm2014} is described in more detail. The simulation design and framework are then presented in Section~\ref{sec:Design}, while Section~\ref{sec:Results} summarizes the main simulation results. In Section~\ref{sec:Example}, a small illustrative real-world data example is shown before the manuscript concludes with a discussion of our findings and an outlook for future research (Section~\ref{sec:Conclusion}).

%
 \section{Methods}
 \label{sec:Methods}
%
 In this section, we explain the univariate and multivariate tree-based methods under investigation. We thereby distinguish between three different approaches: (i) univariate ensembles such as Random Forest and Extra Trees, (ii) their multivariate counterpart and (iii) multi-task Extra Trees.

\subsection{Univariate Tree Ensemble Learner}
To explain the univariate methods, assume that one has access to a training data set $\mathcal{D}_n:=\{ (\mathbf{x}_i^\top,y_i)^\top \in \mathbb{R}^{p+1}: i=1,...,n \}$ consisting of realisations of random vectors with real valued metric outcome $Y$ and  $p$-dimensional metric feature vector $\mathbf{X}$.
\paragraph{Random Forest}
A Random Forest is a univariate machine learning method based on
building ensembles of decision trees. It was developed to address predictive shortcomings
of traditional Classification and Regression Trees (CARTs) \citep{breiman2017classification}. Random Forests consist of a large number
of weak decision tree learners, which are grown in parallel to reduce the bias and variance
of the model at the same time \citep{Breiman2001}. For training a Random Forest, $N$
bootstrap samples are drawn from the training dataset. Each bootstrap sample is
then used to grow an unpruned tree. Instead of using all available features in this step,
only a small and fixed number of randomly sampled $m_{try}$ features are selected as split
candidates. In the regression case,  \cite{Breiman2001} proposed to 
use $m_{try} = \lfloor p/3 \rfloor$, which is still the default choice in many software implementations. A split is chosen by the CART-split criterion for regression, i.e.\ by minimizing the sum of squared errors in both child nodes. These steps are then repeated until $B$ such trees are grown, and new data is predicted by taking the mean of all $B$ tree predictions.
The important hyperparameters for the Random Forest are:
\begin{itemize}
    \item $B$ as the number of grown trees. Note that this parameter is usually not tuned since it is known that more trees are better.
    \item The cardinality of the sample of features at every node is $m_{try}$.
    \item The number of observations in each bootstrap sample for each grown tree is $N$, which is usually the number of observations in the training dataset. 
    \item The minimum number of observations that each terminal node should contain (stopping criteria).
\end{itemize}
\paragraph{Extra Trees}
Extra Trees (also called Extremely Randomized Trees) is an ensemble method developed
by  \cite{Geurts2006} that aims at obtaining trees that are more decorrelated than in the Random Forest approach. Similar to Random Forest, 
Extra Trees consist of $B$ trees, which are also grown in parallel. However, in
contrast to Random Forest, the whole training dataset is used for constructing a tree. For the determination of the next split a random sample of features of size $m_{try}$ is selected. Instead of computing the locally optimal cut-point for each feature in the random sample based on the CART-split criterion, random cut-points are selected. Then, of all the randomly generated splits, the split that optimizes the split criterion is chosen to split the node. The idea of this extra level of randomness is to improve the process of decorrelation leading to smaller variance and a potentially better predictive accuracy in some situations. The important hyperparameters for the Extra Trees are:
\begin{itemize}
    \item $B$ as the number of grown trees.
    \item The cardinality of the sample of features at every node is $m_{try}$.
    \item The number of random split values at each node.
    \item The minimum number of observations that each terminal node should contain (stopping criteria).
\end{itemize}

\subsection{Multivariate Tree Ensemble Learner}
For the multivariate case, we similarly assume that one has access to a training data set $\mathcal{D}_n:=\{ (\mathbf{x}_i^\top,\mathbf{y}_i^\top)^\top \in \mathbb{R}^{p+d}: i=1,...,n \}$ which now consists of realisations of  random vectors with $d$-dimensional metric outcome $\mathbb{Y}$ and $p$-dimensional $\mathbf{X}$ feature vector. To obtain the multivariate versions of Random Forests and Extra Trees, we need to modify the splitting criterion. Here, a natural extension of the CART splitting criteria to multiple metric outputs is to work with the multivariate L2-distance to measure the impurity of a node \citep{Segal2011}. Hence, the impurity function $i$ at node $t$ is given by 
$$
    i(t)=\sum_{\textbf{y}_j\in t} (\textbf{y}_j-\overline{\textbf{y}}(t))\top (\textbf{y}_j- \overline{\textbf{y}}(t)),
$$
where $\overline{\textbf{y}}(t)$ is the sample mean of the output vector at node $t$, see \citep{Death2002} for details. 
Thus, to construct a multivariate Random Forest, we simply proceed as in the univariate case and only change the splitting criterion to the multivariate CART-split extension. The same holds for a multivariate Extra Tree, where for each split, the multivariate impurity function is used to select the optimal split among all randomly chosen splits.

\subsection{Multi-Task Extra Trees}
Another extension of the Extra Trees algorithm was proposed by \cite{Simm2014} in order to handle multi-task learning. For multi-task learning, suppose we have given $T$ supervised learning tasks and all data for the tasks come from the same space ${\bf X} \times Y$, where  ${\bf X} \subset\mathbb{R}^p$ and $Y\subset \mathbb{R}$. Moreover, for each task $t$ we have access to a training data set $$\mathcal{D}_{t,n}:= \left\{({\bf x}^\top_ {i,t}, y_{i,t})^\top \in \mathbb{R}^{p+1} : i = 1,\ldots n \right\},$$
which is sampled from a distribution $P_t$  on ${\bf X} \times Y$. We assume that the $P_t$ are different for each task, but related. \cite{Simm2014}  modified the split-criterion for Extra Trees in such a way that samples can now also be additionally split according to their tasks. These new splits will then create two child nodes and each node contains samples corresponding to separated task subsets. More precisely, when optimizing the split criterion for each node, not only the random feature splits are considered but also a random task split, which is determined as follows:
In the regression case, for each task $t$, the task feature $f_t$ are computed by
$$f_t= \frac{\sum_{v \in I_t}y_v+\alpha \frac{1}{\left|I\right|}\sum_{w\in I}y_w}{\left|I_t\right|+\alpha},$$
 where $I$ denotes the set of sample indexes at the current node, $I_t$ the set of sample indexes of the task $t$ at the current node and $\alpha$ is a regularization parameter weighting the influence of the tasks with default value 1.
Then, the task cut point is randomly selected from the intervall $(\min_t f_t, \max_t f_t).$ This modification is implemented in the R-package \texttt{extraTrees} \citep{Simm2014}.\\

To predict multivariate outputs with the multi-task Extra Tree algorithm, the data has to be transformed as follows: since every component of the multivariate output is considered as a task, the training sample of a multivariate regression problem $$\left\{( {\bf x}_i^\top, {\bf y}_i^\top)^\top\in \mathbb{R}^{p+d}: i = 1,\ldots,n\right\}, $$ is transformed to $$\left\{({\bf x}_i^\top, y_i^j)\in \mathbb{R}^{p+1}: i=1,\ldots, n,\ j = 1,\ldots,d\right\},$$
where ${\bf y}_i=(y_i^1,\ldots y_i^d)^\top.$
The multi-task implementation also requires the following task vector $$\textbf{task}=(1,\ldots, 1,\ldots, d,
\ldots, d)^\top \in \mathbb{R}^{nd},$$ indicating to which component/task the output value $y_i^j$ belongs.
We note, however, that this multivariate extension is only feasible as long as the outputs
are consumerate, i.e. measured on the same scale.
%
\section{Simulation Set-up}
\label{sec:Design}
%
To give an answer to our central research question `when multivariate tree ensemble approaches should be preferred over separate univariate ones' we compare the following machine learning approaches:
\begin{enumerate}
    \item Univariate Random Forests, where for each component of the output vector a 
    separate univariate Random Forest is built.
    \item Univariate Extra Trees, where for each component of the output vector a 
    separate univariate Extra Trees model is constructed.
    \item Multivariate Random Forests based on the extended impurity function $i$.
    \item Multivariate Extra Trees based on the extended impurity function $i$.
    \item Multi-task Extra Trees as proposed in \cite{Simm2014}.
\end{enumerate}
In  extensive simulations we compare these methods with respect to 
(i) predictive power and (ii) runtime. All simulations were conducted in the statistical computing software \texttt{R} \citep{R}. For the Multi-task Extra Trees approach we use the \texttt{extraTrees} package \citep{Simm2014}. For all other approaches, we implemented our own tree construction algorithm to allow a fair runtime comparison among them. The concrete simulation settings are described below. \\

Following \cite{Loh2002, Hsiao2007}, we consider a $3$-dimensional output vector $\mathbf{Y}$ together with $10$ real-valued features $X_1,\ldots, X_{10}$ for which we specifiy different distributions, dependencies and underlying models:

\paragraph{Feature Dependencies and Distributions} Here, in addition to \cite{Loh2002, Hsiao2007}, $X_6,\ldots X_{10}$ are iid  and independent of $X_1,\ldots X_5$. Furthermore, three different dependence structures among the features $X_1,\ldots, X_5$ are considered as summarized in Table~\ref{tab:festures}. In particular, in the first setting (Independent) we consider completely independent features, where $X_1=Z\sim N(0,1)$ is standard normally distributed, $X_2=W\sim exp(1)$ is standard exponentially distributed, $X_3=T\sim t_2$ is $t$-distributed with $df=2$ degrees of freedom, while $X_4=C_4$ and $X_5=C_8$ are uniformly distributed on
$[0,4]$ and $[0,8]$, respectively. This covers symmetric as well as skewed and heavy-tailed distributions. In the other two settings, we modeled a weak dependence between the first three features and a strong dependence between the first two, respectively, see the last two coloumns of Table~\ref{tab:festures}.\\

\begin{table}[h!]
    \centering
   \begin{tabular}{cccc}
\hline
Features&Independent (ind) &Weakly dependent (wd) & Strongly dependent (sd)\\\hline 
$X_1$&$Z$&$Z+W+T$&$W+0.1Z$\\
$X_2$&$W$&$W$&$W$\\
$X_3$&$T$&$T$&$T$\\
$X_4$&$C_4$&$C_2$&$C_2$\\
$X_5$&$C_8$&$C_8$&$C_8$\\
\hline\\
\end{tabular}
    \caption{Distributions of $X_1,\ldots, X_5$ used in the simulation studies. $C_m$ denotes a random variable that is uniformly distributed on $[0,m]$, $T\sim t_2$,  $W\sim\exp(1)$, $Z\sim \mathcal{N}(0,1)$ and $C_m, T, W$ and $Z$ are mutually independent. }
    \label{tab:festures}
\end{table}

\paragraph{Models} Various relationships between the output and some of the features are considered as given in Table~\ref{tab:depresponsefeatures}. The first six dependent models are designed similarly to those in \cite{Loh2002, Hsiao2007}. In addition, we also consider the MGAM and linear models.
Thereby the error $\varepsilon\sim \mathcal{N}(0,\Sigma_{\ell\rho})$ is generated from a multivariate normal distribution with covariance matrix $\Sigma_{\ell\rho}=\left(\begin{array}{rrr}
1&\rho&\rho^\ell\\
\rho&1&\rho\\
\rho^\ell&\rho&1\\
\end{array}\right)$. Here, we distinguish between independence ($\rho=0$) and moderate respectively strong correlations $\rho \in \{0.5, 0.9\}$ of adjacent components. The correlation between the first and last component is either equal to $\rho$ ($\ell=1$) or smaller ($\ell=2$) corresponding to a compound symmetry and autoregressive covariance structure, respectively.
 \begin{table}[h!]
    \centering
    \begin{tabular}{ll}
Jump& $\textbf{Y}=(U+0.7\mathds{1}(X_3>1)\textbf{1}+\varepsilon$\\
Quadratic&$\textbf{Y}=0.8X_2^2\textbf{1}+\varepsilon$\\
Cubic&$\textbf{Y}=0.02X_2^3\textbf{1}+\varepsilon$\\
Additive&$\textbf{Y}=0.7\mathds{1}(X_3>1)\textbf{1}+0.125\sum_{i=1}^8i\mathds{1}(i-1\leq X_5<i)\textbf{1}+\varepsilon$\\
Cross&$\textbf{Y}=0.5\text{sgn}(X_3-1)X_2\textbf{1}+\varepsilon$\\
Random jump (rjump)&$\textbf{Y}=\text{sgn}(X_3-1)U\textbf{1}+
\varepsilon$\\
Linear 1&$\textbf{Y}=\sum_{i=1}^5 X_i\textbf{1}+\varepsilon$\\
Linear 2 &$\textbf{Y}=\sum_{i=1}^{10} X_i\textbf{1}+\varepsilon$\\
MGAM1& $\textbf{Y}= (X_1^2+ \log(X_2)+ \cos(X_3))\textbf{1}+\varepsilon$\\
MGAM2& $\textbf{Y}= (0.1 \sin(X_1), 0.5\log (X_2), U)+\varepsilon$\\
MGAM3 & $\textbf{Y}= (0.1 \sin(X_1), 0.5\log (X_2), U)+
\sum_{i=4}^{10} X_i\textbf{1} + \varepsilon$\\
\end{tabular}
    \caption{Different dependent models between the output and some of the features. $U$ is a uniform random variable on $[0,1]$, $\textbf{1}=(1, 1, 1)^\top$. }
    \label{tab:depresponsefeatures}
\end{table}
For each setting, we generated samples of size $n$ from the respective model with $n \in \{100, 200, 500\}.$ In total, this results in $11$(models)$\times5$(errors)$\times3$(feature dependencies)$\times3$(sample sizes)$=495$ different simulation settings for each of the five ensemble approaches.
\paragraph{Choice of Ensemble Parameters} In order not to have to discuss the different possibilities for hyperparameter tuning, we use the default values recommended in the literature \citep{Breiman2001, Ranger,Hastie2001}. This has the additional advantage of a reduced runtime. Thus, each ensemble learner consists of $500$ trees, the inner bootstrap sample is equal to $m_{try}=3=\lfloor \tfrac{10}{3}\rfloor$, the number of sample points $N$ in the bagging step is equal to the number of sample size $n$. Each terminal node should at least contain five observations. Following the default values of the multi-task Extra Tree implementation \citep{Simm2014}, the number of random cuts is set to one. 
\paragraph{Performance Measures} As the three components of the output are computed on the same scale, we use the overall MSE for all three output components together to evaluate the predictive power. 
To obtain the overall MSE for one setting we used $5-fold$ cross-validation and repeated it $1,000$ times.
Additionally, we also consider the runtime of the algorithms. For each setting we repeated the runtime measurement $1,000$ times. 

%
\section{Results}
\label{sec:Results}
%
In this section, we describe the results of the simulation study. In particular, we present the overall MSE and the runtime of the different construction algorithms under various simulation configurations.

\subsection{Predictive Power}
For ease of presentation, we aggregated the overall MSE  with respect to the $1,000$ replications and $5-fold$ cross-validation. Note that the simulation results of the methods in the setup of weakly dependent features and the MGAM 1 relationship are not presented in the graphics. This is because of their relatively poor performance regarding the predictive power. The methods resulted into average MSE values being greater than $100,000$ (see Figure \ref{fig:MGAM1wd} of the Appendix). \\

\begin{figure}[h!]
    \centering
    \includegraphics[width=1\textwidth]{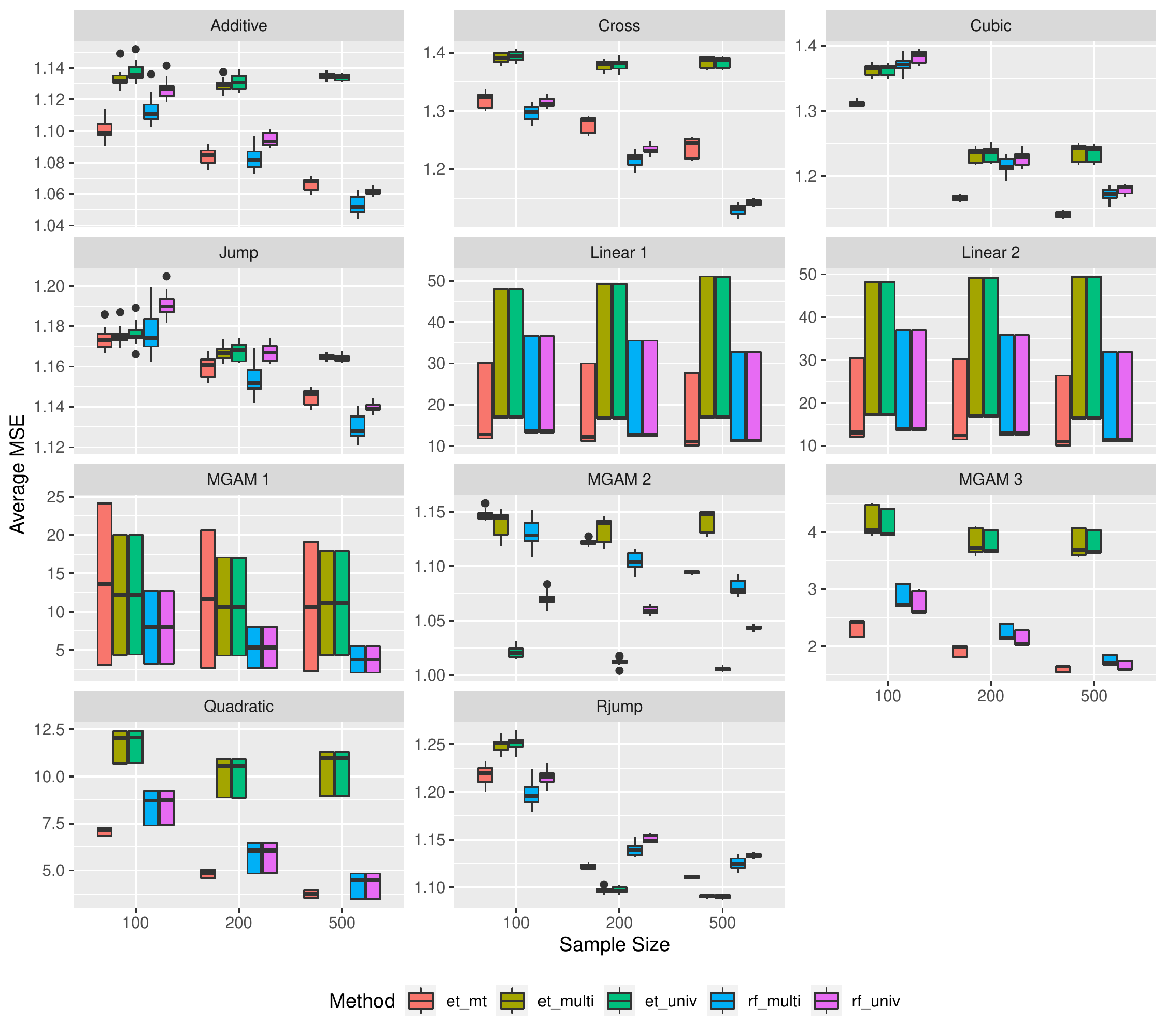}
    \caption{Simulation results on the average MSE for all methods 
multi-task Extra Trees (et\_mt), multivariate Extra Trees (et\_multi), univariate Extra Trees (et\_univ), multivariate Random Forest (rf\_multi) and univariate Random Forest (rf\_univ) separated by the relation between output and feature and sample sizes. }
    \label{fig:MSEn}
\end{figure}

The average MSE for all methods separated by the sample size and the relationship between output vector and features are shown in Figure~\ref{fig:MSEn}.
Here, each boxplot represents $5$(errors)$\times3$(dependencies)$=15$ different average MSE values. Note that, as MGAM 1 with weakly dependent features has been excluded as explained above, the boxplots contain just 10 different average MSE values. Generally, we observe that the performance of the methods depends on the relationship between output and features. Except in the MGAM~2 and random jump setup, either the multivariate Random Forest or the multi-task Extra Trees outperformed the other approaches regarding the predictive power. 
In the MGAM~2 setup, the two univariate approaches have the smallest MSE values. Similar results can be observed for MGAM~3, where the respective differences between univariate and multivariate approaches are small. In all other setups the two Random Forest approaches performed similarly (linear~1, linear~2, MGAM~1 and quadratic) or the multivariate outperformed the univariate approach (additive, cross, cubic, jump, rjump). 
For the Extra Trees, on the other hand, no major differences in the performance of the univariate and multivariate approaches are noticeable except in the case of MGAM2. As the sample size increases, the average MSEs for the two Random Forest approaches decrease in all setups. However, the improvement in the prediction power becomes smaller with increasing sample size. The same observation can be made for the multivariate Extra Trees, but it is different for the multivariate and univariate Extra Trees. While the average MSEs decrease when the sample size is doubled from 100 to 200, the changes in average MSEs are either marginal or the average MSEs increase slightly when comparing sample sizes 200 and 500.\\

\begin{figure}[h]
    \centering
    \includegraphics[width=1\textwidth]{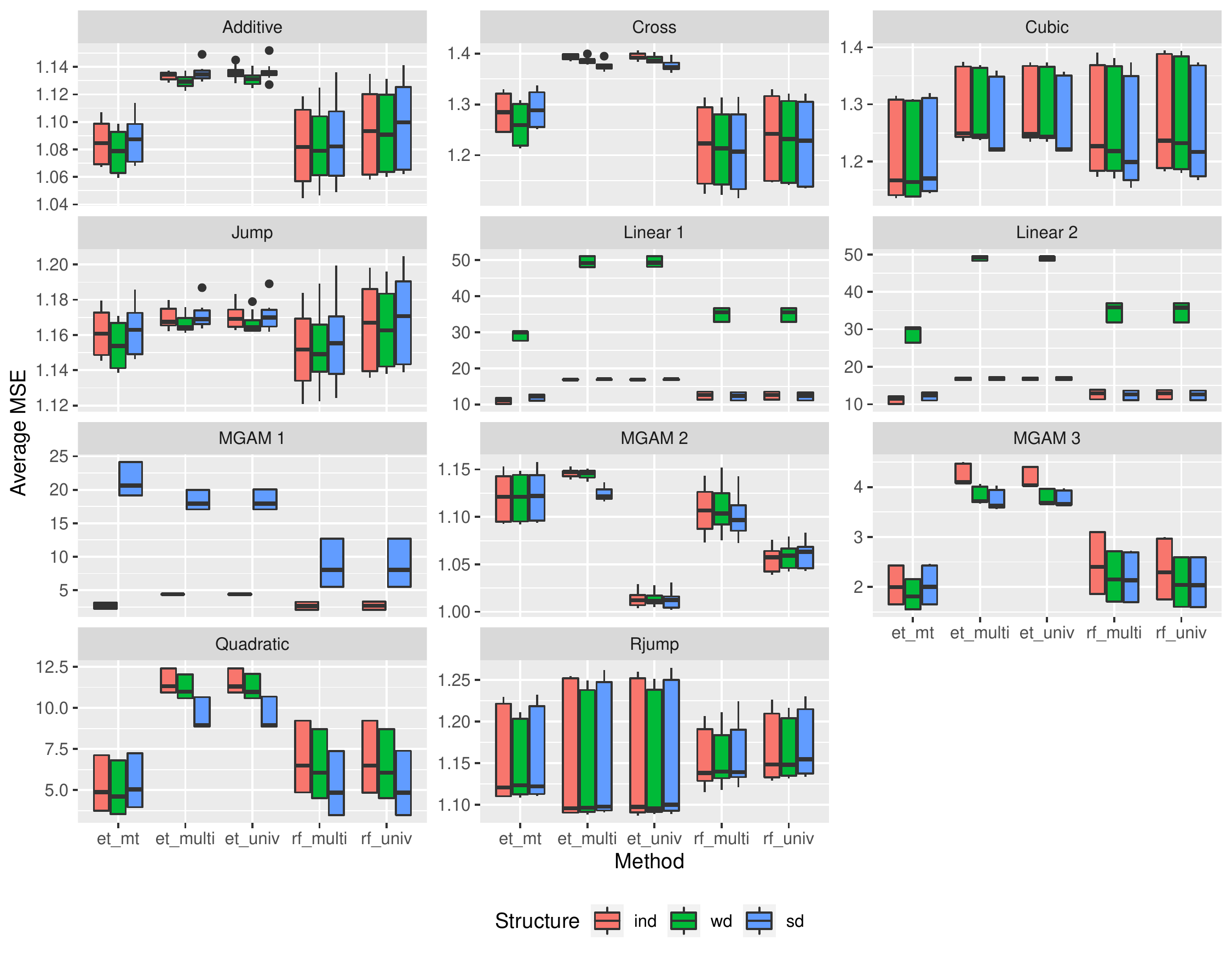}
    \caption{Simulation results on the average MSE for dependency structure of the features separated by the relation between output and feature and all methods multi-task Extra Trees (et\_mt), multivariate Extra Trees (et\_multi), univariate Extra Trees (et\_univ), multivariate Random Forest (rf\_multi) and univariate Random Forest (rf\_univ). Each boxplot represents 5(errors)$\times$3(sample sizes)=15 different average MSE values.}
    \label{fig:MSE_X}
\end{figure}

Figure \ref{fig:MSE_X} summarizes the predicition results for all methods separated by the the relationship between output vector and features and the dependency structure of the features. 
The prediction power of the weakly dependent features of the MGAM 1 setup is shown in Figure~\ref{fig:MGAM1wd} of the Appendix. When the features are weakly dependent, all methods perform poorly in the linear 1, linear 2 and MGAM 1 setup. Overall, the average MSEs are more than twice as large as for the other two feature dependence structures. Except for these settings, the feature dependency structure had only a slight effect on the prediction power. In the MGAM 2 setup, it is noticeable that the average MSEs of the multivariate approaches decrease slightly with increasing feature dependence structure, but the average MSEs of the univariate methods increase. In all other setups, the multivariate and univariate approaches behave the same with increasing dependence structure in the features.\\

\begin{figure}[h!]
    \centering
    \includegraphics[width=1.0\textwidth]{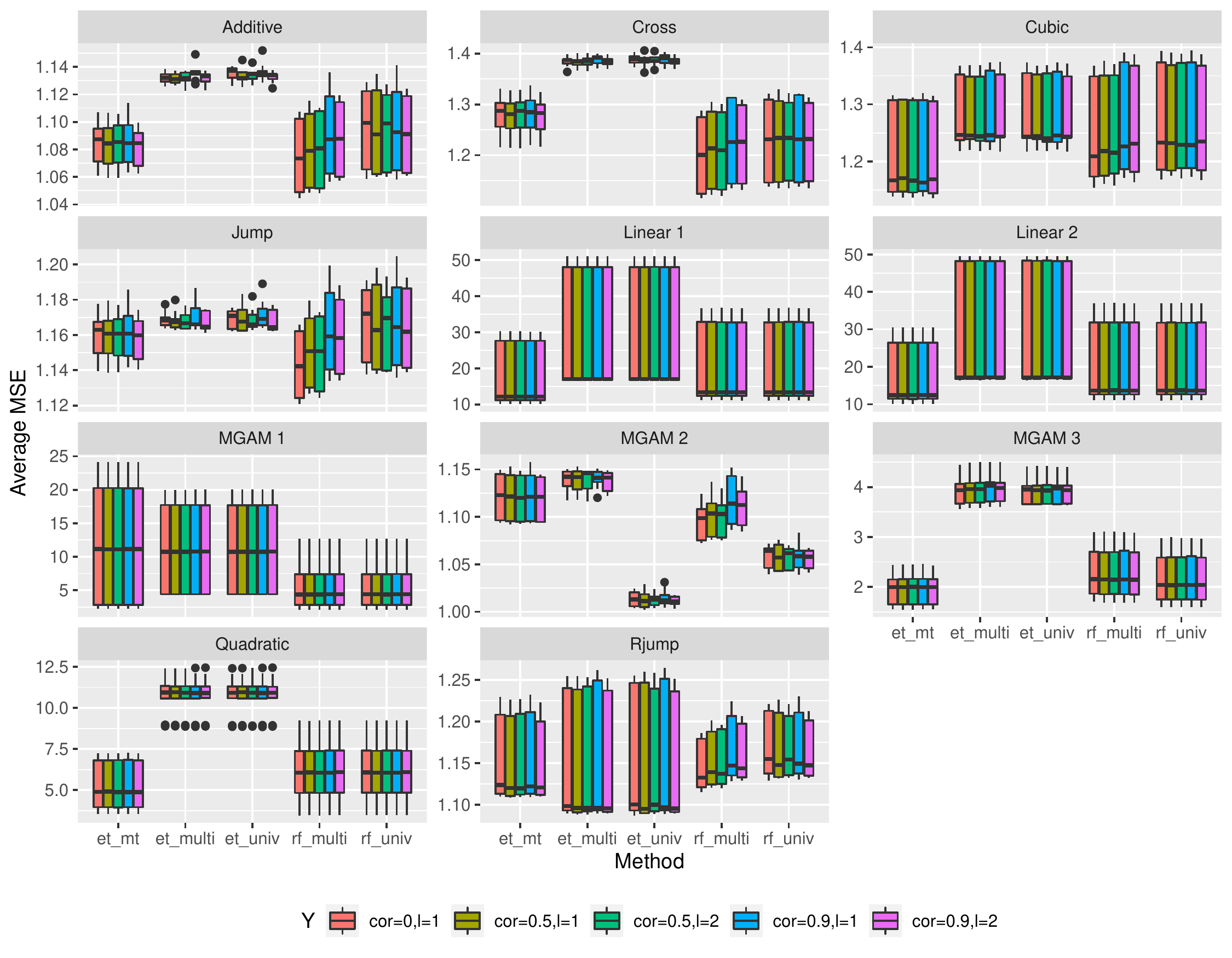}
    \caption{Simulation results on the average MSE for dependency structure of the outputs ated by the relation between output and feature and for all methods multi-task Extra Trees (et\_mt), multivariate Extra Trees (et\_multi), univariate Extra Trees (et\_univ), multivariate Random Forest (rf\_multi) and univariate Random Forest (rf\_univ). The boxplots in the MGAM 1 setup represent 6 different average MSE values and all other boxplots represent 3(feature dependencies)$\times$3(sample sizes)=9 different average MSE values. }
    \label{fig:MSEY}
\end{figure}

The influence of the different errors on the prediction power can be seen in Figure~\ref{fig:MSEY}. 
Generally, the different errors had little impact on the prediction power. For setting linear 1, linear 2, MGAM~1, MGAM~3 and quadratic are no differences in the average MSE values between the errors for all methods. In all other setups, the average MSEs of the multivariate Random Forest increase with an increasing dependency structure in the outputs.

\subsection{Runtime}
The runtimes for all different setups are given in Figures~\ref{fig:RuntimeX} and \ref{fig:RuntimeY} of the Appendix. 
Studying the results in detail, we realize that the dependency structures of the features and the dependency structures of the output had little impact on the runtime. For ease of presentation, we therefore aggregated the runtime of the multiple simulation settings with regard to both dependency structures and the number of repetitions in Table~\ref{tab:AverageRuntime}. It summarizes the mean runtime for each construction method separated by the sample sizes and the relationship between output and features.
\begingroup
\fontsize{8}{14}\selectfont
\setlength{\tabcolsep}{1.9pt} 
\begin{table}[h!]
\resizebox{\textwidth}{!}{%
\begin{tabular}{l@{\hspace{8pt}}ccccc@{\hspace{8pt}}ccccc@{\hspace{8pt}}ccccc}
\toprule
 & \multicolumn{5}{c}{$n=100$\hbox{\hspace{9pt}}} & \multicolumn{5}{c}{$n=200$\hbox{\hspace{9pt}}} & \multicolumn{5}{c}{$n=500$\hbox{\hspace{9pt}}} \\
 \cmidrule(l{1.9pt}r{11pt}){2-6} \cmidrule(l{1.9pt}r{11pt}){7-11} \cmidrule(l{1.9pt}r{11pt}){12-16} 
Setting &et\_mt&et\_multi&et\_univ&rf\_multi&rf\_univ&et\_mt&et\_multi&et\_univ&rf\_multi&rf\_univ& et\_mt & et\_multi&et\_univ&rf\_multi&rf\_univ \\
 \midrule
Linear 1 & 0.09 & 35.12 & 90.72 & 197.48 & 351.47 & 0.19 & 57.51 & 148.44 & 466.85 & 799.12 & 0.52 & 94.91 & 242.25 & 1464.91 & 2397.94 \\
Linear 2 & 0.09 & 35.59 & 92.35 & 198.22 & 352.56 & 0.19 & 61.49 & 158.59 & 480.20 & 822.13 & 0.52 & 112.06 & 286.17 & 1485.69 & 2435.09 \\
Additive & 0.09 & 24.48 & 51.11 & 205.52 & 365.07 & 0.19 & 21.68 & 49.51 & 501.73 & 854.92 & 0.52 & 11.56 & 28.56 & 1511.66 & 2476.85 \\
Cross & 0.10 & 35.07 & 73.51 & 200.93 & 354.34 & 0.20 & 45.65 & 92.20 & 496.67 & 838.58 & 0.53 & 28.47 & 67.30 & 1574.91 & 2549.15 \\
Cubic & 0.10 & 34.59 & 70.13 & 202.80 & 354.38 & 0.20 & 43.17 & 83.76 & 512.42 & 852.50 & 0.53 & 21.75 & 53.67 & 1697.36 & 2709.18\\
Jump 1& 0.09 & 34.40 & 69.35 & 199.73 & 352.07 & 0.19 & 40.80 & 78.93& 494.31 & 835.34 & 0.53 & 15.27 & 40.26& 1603.17 & 2602.08 \\
MGAM 1 & 0.09 & 31.47 & 79.42 & 207.01 & 361.96 & 0.20 & 42.71 & 106.84& 515.78 & 862.85 & 0.53 & 40.20 & 99.86 & 1661.02 & 2639.98 \\
MGAM 2 & 0.09 & 37.89 & 71.30 & 197.38 & 351.83 & 0.20 & 51.81 & 85.44 & 494.44 & 846.71 & 0.55 & 30.39 & 49.63 & 1541.62 & 2594.42 \\
MGAM 3 & 0.09 & 35.77 & 86.17 & 185.40 & 332.39 & 0.19 & 62.06 & 145.50 & 447.06 & 773.65 & 0.51 & 117.51 & 264.01 & 1389.48 & 2302.53 \\
Quadratic &0.09 & 31.98 & 76.54 & 204.49 & 356.83 & 0.19 & 42.24 & 98.86 & 509.71 & 851.53 & 0.52 & 33.67 & 79.25 & 1625.67 & 2595.54 \\
Rjump &0.09 & 34.15 & 69.65 & 197.58 & 348.84 & 0.20 & 40.77 & 80.19 & 486.75 & 822.54 & 0.53 & 16.09 & 41.99 & 1572.21 & 2545.54 \\
\bottomrule

\end{tabular}
}\caption{Average runtimes (in seconds) for the dependency models between the output and the features, aggregated over all dependency settings of the outputs and features and all repetitions for the methods 
multi-task Extra Trees (et\_mt), multivariate Extra Trees (et\_multi), univariate Extra Trees (et\_univ), multivariate Random Forest (rf\_multi) and univariate Random Forest (rf\_univ).
}
\label{tab:AverageRuntime}
\end{table}
\endgroup
Since the multi-task method was programmed runtime-efficiently in Java \citep{Simm2014}, unlike the other methods, this method is the fastest. Its average runtimes are faster than 0.6 seconds for all settings. In addition, the runtime behaves approximately linear to the sample size with about 0.1 seconds for $n=100$ up to $0.6$ seconds for $n=500$. \\

Considering our implementations, we first note that the Extra Trees algorithms are faster than the Random Forest algorithms. 
 The multivariate Random Forest requires on average (overall settings and sample sizes) 25 times as long as the multivariate Extra Trees. In the univariate case, the Random Forest takes on average 18 times as long as the Extra Trees algorithm. This time difference is probably due to the different choices of split values in the methods. Extra Trees randomly selects the split values, while Random Forest uses an exhaustive search to find these values.
More important for us is a comparison between the multivariate and the univariate approaches: the results show that in both cases the multivariate approaches require less runtime than the univariate approaches. In fact, the multivariate Extra Trees show a decrease in runtime between 38\% and 62\% compared to the univariate approach.  Using the multivariate Random Forest approach decreases runtime by 37\% to 44\% compared to the corresponding univariate method. For both Random Forest approaches, average runtimes increase similarly with increasing sample size. When sample size doubles from 100 to 200, runtimes for both approaches increase by 127-153\%, while runtimes increase by 190-231\% when sample size changes from 200 to 500. The different relations between output and features have a small effect on runtimes, as runtimes for fixed sample sizes (compared to the fastest setting MGAM~3) increase by 6-22\% for multivariate Random Forests and 3-17\% for univariate Random Forests.\\

\begin{figure}
    \centering
    \includegraphics[width=0.8\textwidth]{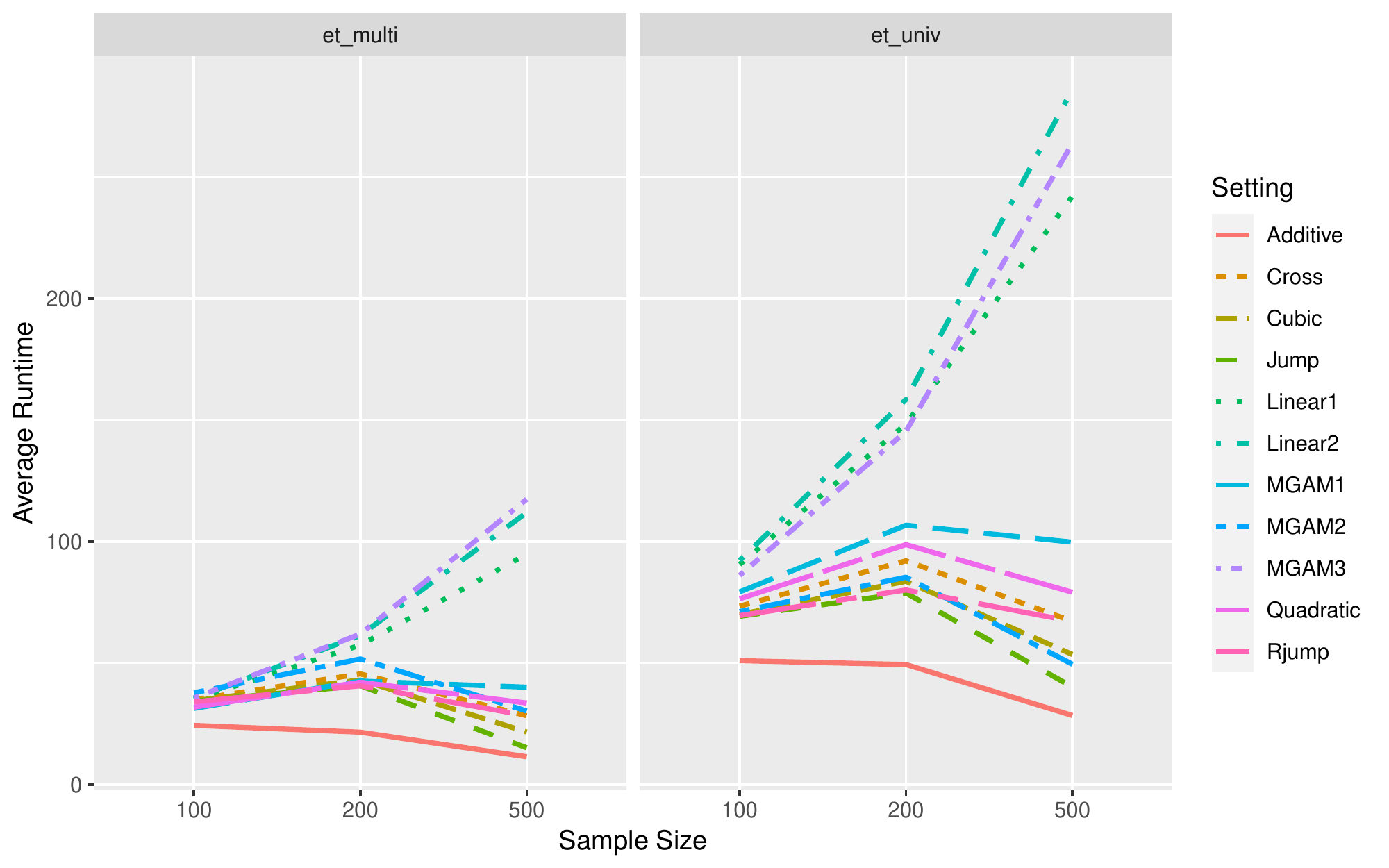}
    \caption{Average runtimes for multivariate (left) and univariate Extra Trees (right) aggregated over all dependency settings of the outputs and features and all repetitions.}
    \label{fig:etruntimes}
\end{figure}

A different observation can be made for the multivariate and univariate Extra Trees. When the sample sizes are increased, their runtime behaviour varies from setting to setting, see Figure~\ref{fig:etruntimes}.
For the settings linear 1, linear 2 and MGAM 3, the runtimes increase for increasing sample sizes. Note that these are also the settings with the largest runtimes for a fixed sample size. However, the setting additive has the fastest runtimes and decreases slightly with increasing sample size. The runtimes of the other settings increase by 13.81-36.75\% when the sample size is doubled from 100 to 200,  but decrease by 5.87-62.56\% when increasing the sample size from 200 to 500.

%
\section{Illustrative Real-World Data Example}
\label{sec:Example}
\begin{figure}[h!]
    \centering
    \includegraphics[scale=0.75]{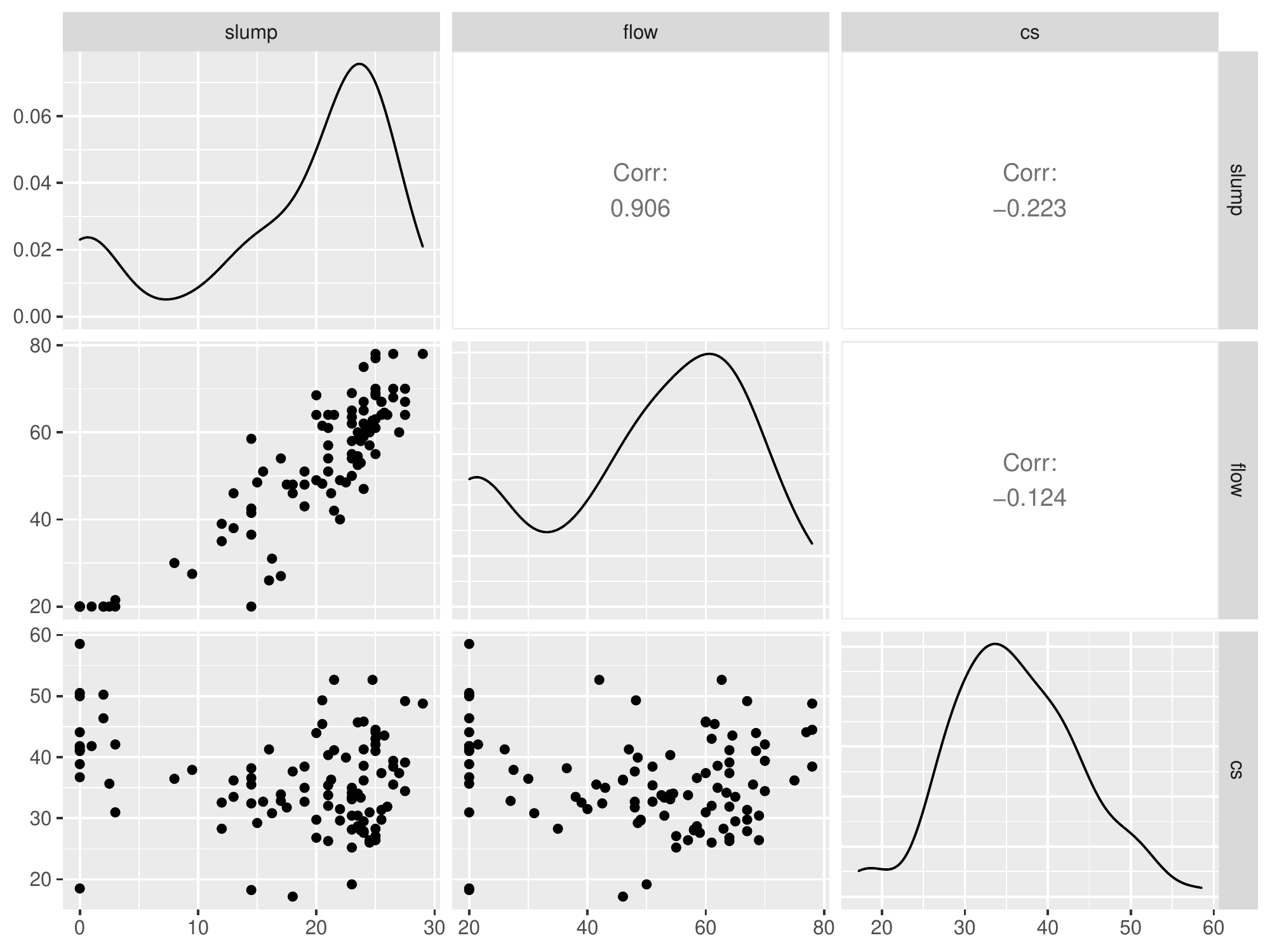}
    \caption{Plots of the output vector (using the unscaled data): Empirical Pearson correlation coefficients (upper triangle), density plots (diagonal) and scatterplots (lower triangle).}
    \label{fig:DataConcrete}
\end{figure}

%
We consider the concrete slump test study \citep{Yeh2007} taken from the UCI Machine Learning Repository \citep{Dua2019}. The dataset consists of 103 complete observations on seven continuous features (all are ingredients of concrete and measured in kg$/$m$^3$): cement, slag, fly ash, water, superplasticizer, coarse aggregate and fine aggregate and three continuous output variables: slump (in cm), flow (in cm) and 28-day compressive strength (cs; in Mpa). Slump is the vertical height by which a cone of wet concrete sinks while
flow is the horizontal distance by which the concrete spreads. Both variables are used to measure the viscosity of
concrete.
The dependencies within the output vector are shown in Figure~\ref{fig:DataConcrete}. In particular, slump and flow are highly positively correlated and the correlation between cs and slump/flow is slightly negative. In the following analyses, the outputs were standardized to zero mean and unit variance.\\

We now investigate our central research question `when multivariate tree ensembles approaches should be preferred over separate univariate ones’ for this specific dataset. Therefore, we construct the trees for different dimensions of outputs (uni-, bi- and trivariate) using all the methods presented in Section~\ref{sec:Methods}.
To compare the overall MSE of the approaches, we applied  $5-fold$ cross-validation and repeated it $100$ times. As reported in Table~\ref{tab:MSEconcrete}, the RF approaches have the smallest MSE values (between $0.347$ and $0.657$) in all output dimension settings. 
When comparing the univariate and multivariate approaches, both RF approaches show almost similar results with a slight advantage (at the second or third decimal position) for the univariate RF.
For the Extra Trees the observation is vice versa, i.e. multivariate Extra Trees lead to a slightly lower MSEs than the univariate ones in all multivariate settings.

\begingroup
\centering
\fontsize{8}{12}\selectfont
\setlength{\tabcolsep}{7pt} 
\begin{table}[h!]
\resizebox{0.95\textwidth}{!}{%
   \hspace{25pt}\begin{tabular}{l@{\hspace{13pt}}ccc@{\hspace{13pt}}ccc@{\hspace{13pt}}c}
\toprule
 & \multicolumn{3}{c}{univariate\hbox{\hspace{9pt}}} & \multicolumn{3}{c}{bivariate\hbox{\hspace{9pt}}} & \multicolumn{1}{c}{trivariate\hbox{\hspace{9pt}}} \\
 \cmidrule(l{1.9pt}r{11pt}){2-4} \cmidrule(l{1.9pt}r{11pt}){5-7} \cmidrule(l{1.9pt}r{11pt}){8-8} 
Method&slump& flow&cs&slump and flow&slump and cs&flow and cs &slump, flow and cs\\
 \midrule
RF uni& 0.657&0.583&0.347&0.620&0.502&0.465&0.529\\
RF multi&0.657&0.583&0.347&0.619&0.523&0.497&0.563\\
ET uni&0.809&0.722&0.594&0.766&0.701&0.658&0.708\\
ET multi&0.809&0.722&0.594&0.761&0.674&0.634&0.683\\
ET mt&0.686&0.588&1.469&1.077&1.050&1.039&1.088\\
\bottomrule
\end{tabular} 
}     \caption{Average MSE of the construction methods presented in Section \ref{sec:Methods} using concrete dataset based on 5-fold crossvalidation and 100 repetations.}
\label{tab:MSEconcrete}
\end{table}
\endgroup
To compare the prediction accuracy of our approaches to MVPART \citep{MVPART}, univariate and multivariate GUIDE \citep{Loh2013}, we follow \cite{Loh2013} and apply leave-one-out cross-validation to estimate the sum of MSEs of the trees, where the sum is over the three output variables. In contrast to \cite{Loh2013}, we do not prune the trees in our approaches. The results are shown in Table \ref{tab:CompareLoh}. Note that we include the results of \cite{Loh2013} in the table without performing their experiments ourselves. While the multivariate Extra Trees outperformed their univariate method by decreasing the sum of MSE by 5.941\%, the univariate RF and GUIDE slightly decreased the sum of MSEs by 5.458\% to 7.15\%s compared with their multivariate approaches. In general, the Random Forest approaches have the smallest sum of MSEs (1.594/1.681).
\begin{table}[h!]
    \centering
    \begin{tabular}{ccccccccc}
    \hline
       Methods  &  RF uni& RF multi& ET uni& Et multi& ET mt&MVPART& GUIDE uni& GUIDE multi\\\hline
         Sum of MSE&1.594&1.681&2.12&2.02&3.42&2.096&1.957&2.097 \\\hline
    \end{tabular}
    \caption{Sum of average MSEs of methods considered in Section \ref{sec:Methods} and \cite{Loh2013} using concrete data set based on leave-one-out crossvalidation.}
    \label{tab:CompareLoh}
\end{table}

%
\section{Conclusion and Outlook}
\label{sec:Conclusion}
%

The main purpose of this simulation study was to compare the predictive accuracy of univariate Random Forests and Extra Tree algorithms with multivariate approaches in case of multivariate outputs.
\\

In most of the simulation settings it was clearly shown that either the Random Forest or the Extra Tree approaches have yielded better performances than the other methods. However, when comparing the multivariate approaches with their univariate counterparts, then in two of the simulation settings (MGAM 2 and MGAM 3) advantages for the univariate approaches considering the performance (average MSE) could be found. In all other simulation settings the multivariate approaches have shown at least similar or even better performances than the univariate approaches. Especially, when comparing univariate and multivariate Random Forest approaches, in some of the considered settings the performances of the multivariate approaches were substantially better. Moreover,
for all methods, the different dependency structures within the covariates did have an impact on the methods' performance. 
However, it was virtually the same for both, the univariate approaches and the multivariate approaches meaning not one approach had an advantage over the others. 
Another interesting finding was that correlation within the outputs only showed substantial impacts on the performance of the multivariate Random Forests in some simulation settings, while for all other approaches this had little to no impact at all.\\

While the differences of the univariate and multivariate approaches' predictive performance were small to moderate, a huge difference in the runtime could be noted. Here, the multi-task Extra Tree method was the fastest approach. However, it was the only method that was not specially implemented for this simulation study. For a fair runtime comparison the other approaches were implemented in a comparable way and it was shown that the multivariate approaches have a huge runtime advantage over the univariate approaches.\\

Last but not least, with a real data example it was shown that the multivariate approaches can improve the performance when considering multivariate outputs, in this case especially for the Extra Tree approaches. However, for the Random Forest approaches  
the multivariate counterpart could only improve the performance in one of the bivariate cases.\\

As only regression problems with numeric outputs were considered, future simulation studies should investigate whether the same potential for improvement can also be found for multivariate classification problems. Also, mixed problems with numeric and categorical outputs at the same time have to be investigated. Moreover, as in this study only rather low sample sizes were investigated, the behavior of these approaches in big data settings with larger sample sizes should also be further researched.

\bibliography{mybibfile}
          \newpage
   \appendix
    \section{Additional Simulation Results}
    \begin{figure}[h]
        \centering
        \includegraphics[width=1\textwidth]{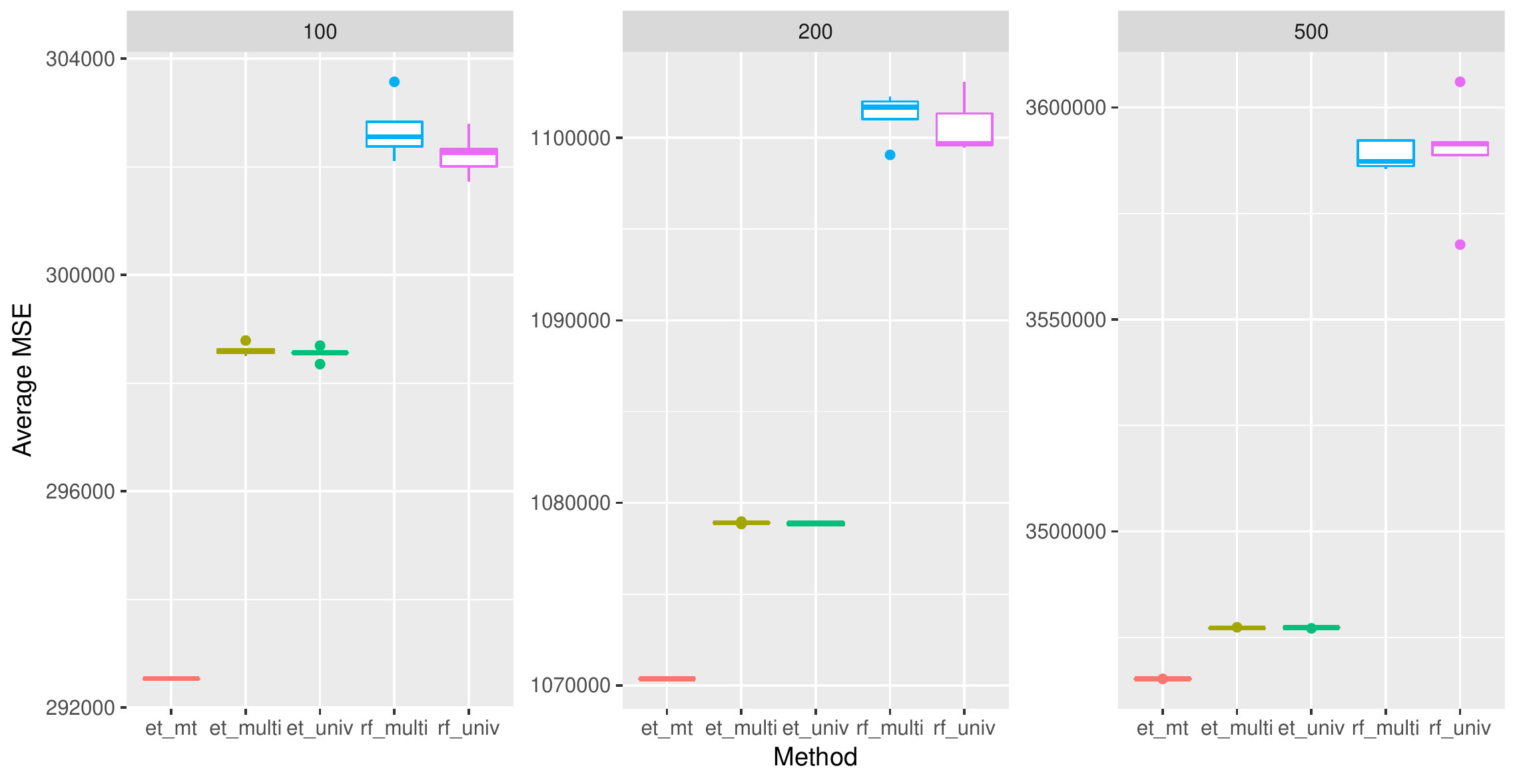}
        \caption{Average MSE for the setup of weakly dependent features and the MGAM 1 relation between features and outputs separated by the sample size and the methods 
multi-task Extra Trees (et\_mt), multivariate Extra Trees (et\_multi), univariate Extra Trees (et\_univ), multivariate Random Forest (rf\_multi) and univariate Random Forest (rf\_univ). Each boxplot represents 5 average MSE values.}
        \label{fig:MGAM1wd}
    \end{figure}
\begin{figure}[h]
    \centering
    \includegraphics[width=\textwidth]{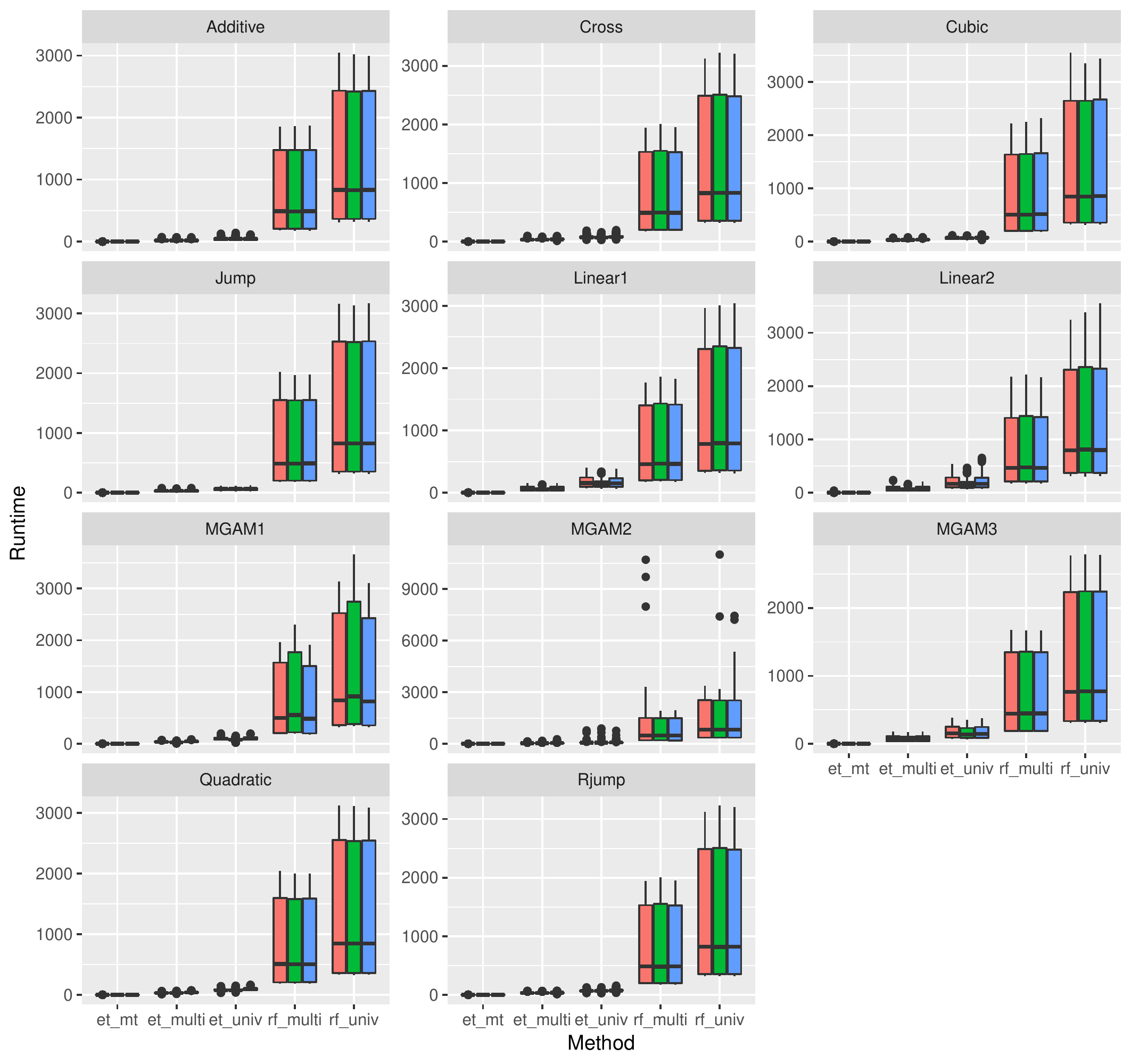}
    \caption{Simulation results on the runtime for the feature dependence settings separated by the relation between output and feature and all methods 
multi-task Extra Trees (et\_mt), multivariate Extra Trees (et\_multi), univariate Extra Trees (et\_univ), multivariate Random Forest (rf\_multi) and univariate Random Forest (rf\_univ). The arranged boxplots in each method segment correspond to the following feature dependence structures: (from left to right) independent (red), weakly dependent (green), and strongly dependent (blue). Each boxplot contains 15,000 runtime measurements.}
    \label{fig:RuntimeX}
\end{figure}
\begin{figure}[h]
    \centering
    \includegraphics[width=\textwidth]{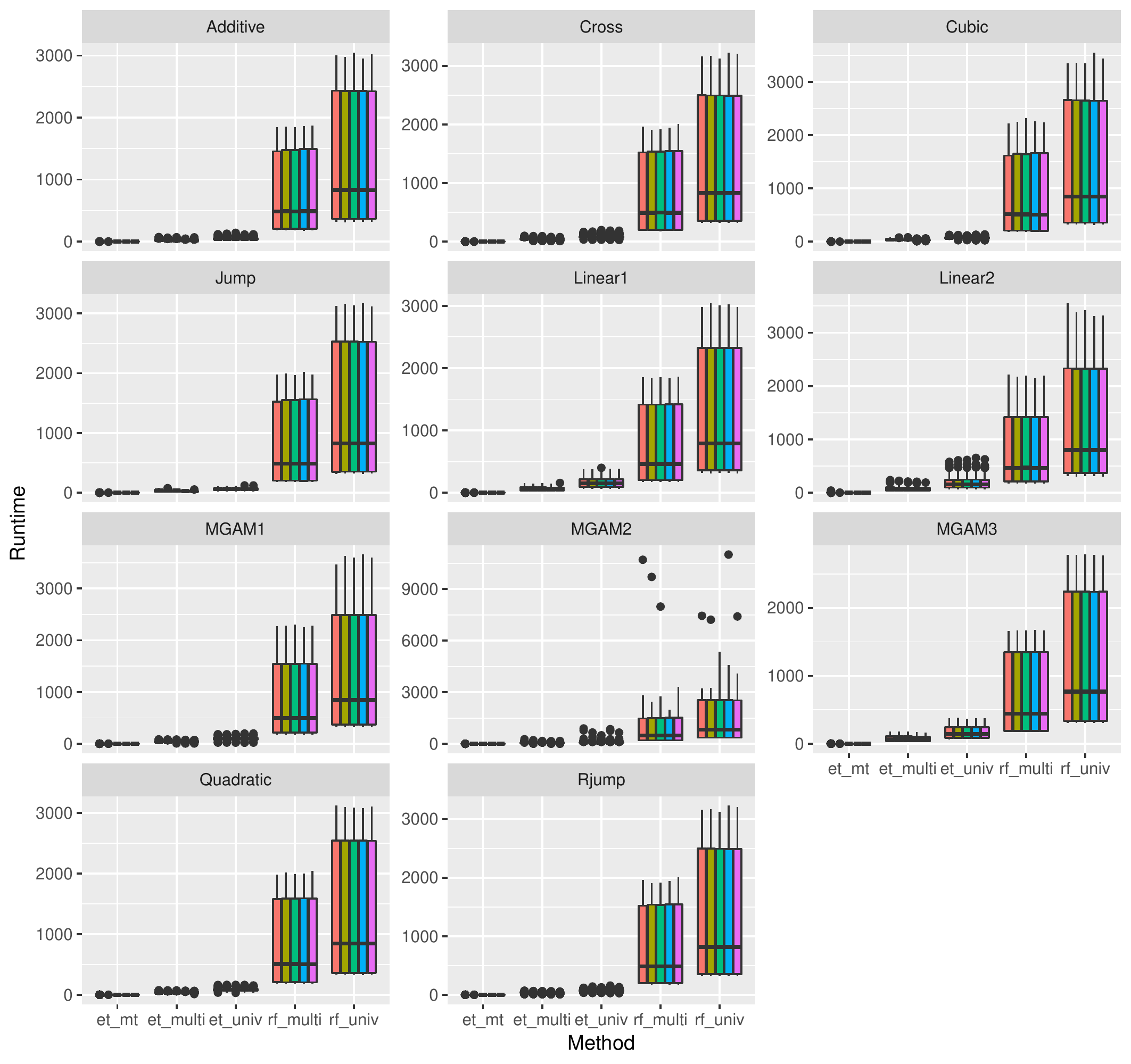}
     \caption{Simulation results on the runtime for the output dependence settings ated by the relation between output and feature and all methods 
multi-task Extra Trees (et\_mt), multivariate Extra Trees (et\_multi), univariate Extra Trees (et\_univ), multivariate Random Forest (rf\_multi) and univariate Random Forest (rf\_univ). The arranged boxplots in each method segment correspond to the following dependence structures: (from left to right) cor=0 and $\ell=1$, cor=0.5 and $\ell=1$, cor=0.5 and $\ell=2$, cor=0.9 and $\ell=1$ and cor=0.9 and $\ell=2$. Each boxplot contains 9,000 runtime measurements.}
    \label{fig:RuntimeY}
\end{figure}

\end{document}